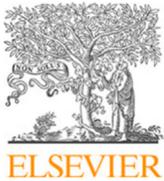
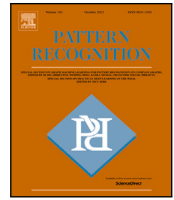

# Enhancing textual textbook question answering with large language models and retrieval augmented generation

Hessa A. Alawwad [a,b,*], Areej Alhothali [a], Usman Naseem [c], Ali Alkhathlan [a], Amani Jamal [a]

[a] *Faculty of Computing and Information Technology, King Abdulaziz University, Jeddah, Saudi Arabia*
[b] *College of Computer and Information Science, Imam Mohammad Ibn Saud Islamic University (IMSIU), Saudi Arabia*
[c] *School of Computing, Macquarie University, Australia*

ABSTRACT

Textbook question answering (TQA) is a challenging task in artificial intelligence due to the complex nature of context needed to answer complex questions. Although previous research has improved the task, there are still some limitations in textual TQA, including weak reasoning and inability to capture contextual information in the lengthy context. We propose a framework (PLRTQA) that incorporates the retrieval augmented generation (RAG) technique to handle the "out- of-domain" scenario where concepts are spread across different lessons, and utilize transfer learning to handle the long context and enhance reasoning abilities. Our architecture outperforms the baseline, achieving an accuracy improvement of 4. 12% in the validation set and 9. 84% in the test set for textual multiple-choice questions. While this paper focuses on solving challenges in the textual TQA, It provides a foundation for future work in multimodal TQA where the visual components are integrated to address more complex educational scenarios. Code: https://github.com/hessaAlawwad/PLR-TQA

## 1. Introduction

Natural language processing (NLP) is one of the most complex applications of artificial intelligence (AI). Using deep learning algorithms to enhance a computer's ability to either understand natural language or generate natural language has contributed to the latest advancements in the NLP field [1]. Question answering (QA) is particularly captivating and expansive among the many tasks in NLP. As a part of AI and NLP, QA employs NLP techniques to respond to queries [2]. This task requires a deep understanding of language to provide accurate answers to questions posed in natural language by humans.

Research in QA has shown significant activity, classifying QA systems into three categories based on input modality or knowledge source [3]: Context QA, visual QA (VQA), and textbook QA (TQA). Context QA, also known as machine reading comprehension (MRC) [1], involves a model answering a natural language question by understanding textual context. VQA integrates NLP and computer vision (CV), requiring a model to deduce answers from images. TQA requires a model to answer multimodal questions by comprehending multimodal contexts [3,4].

In 2017, Kembavi and other researchers [4] proposed a dataset to introduce challenges unique to TQA, which are absent in MRC and VQA. TQA involves complex reasoning and understanding of multimodal contexts, which include scientific diagrams and extensive information, in order to answer questions.

TQA is a complex task that aims to answer questions by comprehending the middle school science education materials that encompass visual and textual content in order to answer diverse types of questions. The complexity of the task comes from the challenges it brought and the required reasoning ability over the integrated information from different modalities, which require both visual comprehension and language understanding.

Considered an AI grand challenge [5], TQA represents a multimodal machine reading comprehension ($M^3C$) task where both modalities have their own challenges and difficulties, demanding substantial research efforts. For the textual part of TQA which is our main focus in this paper, the longer average length of context, where the answer must be inferred from, has an average length of 1800 words [6].

Effectively handling dependencies within such lengthy textual contexts is a crucial challenge. The extensive nature of the context, with more than 50 sentences in over 75% of the lessons [4], combined with the limited training data (15,153 samples), adds complexity to representing the diversity and intricacy found in real-world data, which hinders a model's ability to effectively generalize. Another challenge is that some concepts are explained in different lessons within TQA, which is referred to as "out-of-domain" problem. The out-of-domain problem brings the need for some techniques that retrieve the relevant context from all the available lessons and not only the lesson where the question in.

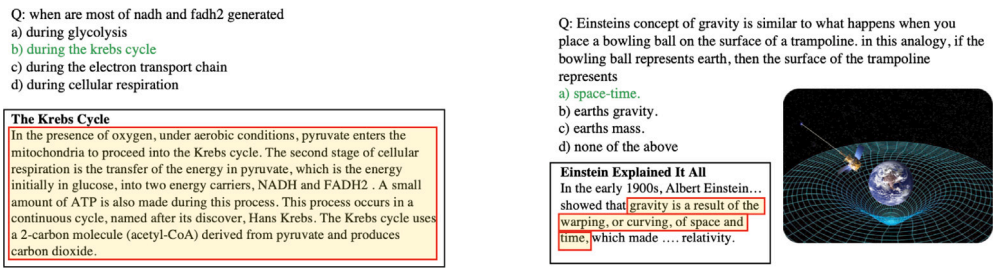

**Fig. 1.** Examples from the textbook question answering dataset [4]. The question on the left requires inferring the answer from two sentences. On the right, images present analogies in the textbook question answering dataset, where the question aims to clarify the notion of gravity by establishing a comparison that highlights similarities between two distinct physical scenarios, making the complex concept more comprehensible.

The TQA dataset (CK12-QA) demonstrates a high level of understanding, as shown in Fig. 1. The question on the left requires extracting an answer from multiple sentences. In contrast, the question on the right aims to compare two scenarios to explain a complex concept in a simpler or more relatable way. It also requires the ability to handle qualitative and quantitative data, along with a deep understanding of language nuances related to negation, conjunction, or common sense [7].

Recent works on TQA used advanced deep learning techniques but still lacks the ability to resolve some of the existing challenges, more specifically the challenge of capturing the textual information on the context and the ability to adequately reason over such lengthy lessons. The TQA task is vital due to its real-life modality, reflecting the complexities of questions and contexts found in textbooks, emphasizing its importance as a knowledge source and the achieved accuracy needs to be improved.

These challenges must be addressed and resolved by incorporating the latest advancements in both MRC and CV. Recent advancements in Large Language Models (LLM) such as, GPT [8] and Llama-2 [9], have demonstrated an improved understanding and comprehending of text and question answering [10]. When information is provided within the LLM context window, these models excel at generating responses that are cohere and contextually relevant. However, LLMs have significant difficulties when used for TQA, especially when dealing with long-context inquiries and out-of-domain scenarios. LLMs can be misleading in these cases where the actual relevant information might be scattered across the different lessons or even missing causing inaccurate answers. We suggest a novel framework that combines the Llama-2 large language model with retrieval-augmented generation (RAG) to address these issues.

Llama-2, an LLM, has achieved the highest performance among open-source LLMs, surpassing models like Falcon [11] on standard academic benchmarks, including common-sense reasoning, world knowledge, and reading comprehension. Despite its straightforward training approach, Llama-2 demonstrates capabilities across various NLP tasks [9]. While large language models exhibit remarkable capabilities in generating and understanding natural language, fine-tuning may be necessary for optimal performance in specific tasks or domains. Another advancement that enhances the ability of an LLM to generate text is the retrieval augmented generation (RAG), which is concerned with augmenting the LLM context window with the most relevant text.

Our approach leverages the strengths of both components: The first component is RAG, which addresses three key challenges in TQA: (1) It effectively handles long lessons by retrieving only the most relevant paragraphs from the textbook, ensuring that information is more focused. (2) It reduces noise that could distract the answer generation process by filtering out irrelevant content. (3) RAG solves the "out-of-domain" problem by retrieving external knowledge from other lessons that may not be present in the immediate lesson context, allowing the model to handle questions with missing or spread-out information. Llama-2, the second component, enhances the model's ability to reason over the retrieved passages and benefits from its extensive pre-training, allowing it to generate answers that are both accurate and contextually relevant, even in challenging scenarios.

Although TQA is a multimodal task by nature, involving text as well as visual components like figures and diagrams, this study concentrates on the textual part of TQA. We believe that improving the textual TQA is an essential first step towards developing a more complete TQA system that can process multimodal inputs. In order to fully address the challenges of the TQA task, future research will investigate the integration of visual data. Our model achieves state-of-the-art performance in both true/false and multiple-choice question answering, showing notable improvements over current approaches. These results demonstrate how much our retrieval-augmented method improves TQA performance.

The paper is organized as follows: An overview of the related works and the dataset used in our investigation is given in Sections 2 and 3, respectively. The architecture and implementation details of the components are explained in Section 4. In Section 5, we present the experiments along with their findings and further extend the experiments in the ablation study. Section 6 finally brings our efforts to a close.

## 2. Related work

Research on TQA started with the main paper MemN [4] where they applied MRC models like BiDAF [12] and VQA models [4]. Those models performed poorly with an accuracy 2.49% less than the lowest validation accuracy achieved and 35.78% less than the highest validation accuracy achieved on the TQA dataset due to the inherent challenges the dataset presents.

Research on TQA has been categorized into three sets according to [13]: graph-based, pretraining-based, and interpretability-based.

The graph-based studies focus on utilizing graph-based methods to address TQA, including the works of MoCA [14], IGMN [6], RAFR [15] and, DIMP [16]. MoCA tried to solve the gap between specific and general domains by introducing pre-trained language models on a general domain. They regard textual context and diagrams as knowledge and select the set of top-k supporting sentences and graph nodes (knowledge) for a question and then compare the performance of different knowledge representations. They apply external knowledge to enhance the span-level representation. The text in MoCA was encoded using a multi-stage pre-trained module (RoBerta) that was pre-trained on Wikipedia and BookCurpus dataset for the Random mask strategy and CK12-QA for the span mask strategy. The encoder was fine-tuned on RACE and TQA datasets. To obtain the features of the question and instructional diagrams in MoCA they used a transformer encoder and added a linear projection layer to align the text and visual parts of the image into a common space. To enhance the model, they used a feedforward and layer normalization to obtain the features from the multihead-guided attention layer.

IGMN attempted to find the contradictions between textual contexts and candidate answers to build a Contradiction Entity-Relationship Graph (CERG). They utilized hand-written semantic rules to comprehend long essays via CERG where they used the Stanford Parser and the Natural Language Toolkit (NLTK) to build it. They utilize spatial analysis rules to comprehend diagrams via contradiction entity-relationship graph. They used BiLSTM and VGG Net.

RAFR seeks to learn an effective diagram representations and the questions, options, and the closest paragraph were fed into an LSTM to get their representations. They analyze the relative positions and dependencies between text within diagrams to build a relation graph based and then apply dual attention to predict answers. They apply graph attention networks (GATs) to understand diagrams. RAFR considers only the text on the diagram, and that causes a loss of other important visual information.

DIMP introduced the Diagram Perception Network, an end-to-end model that integrates diagram understanding with text comprehension. DIMP adopts a joint optimization strategy that simultaneously handles relation prediction, diagram classification, and question answering tasks. The DIMP model uses a graph-based approach to handle diagram understanding by representing diagrams as relational graphs and capturing semantic propagation within them.

The pretraining-based papers propose a multistage pre-training approach for the model, followed by fine-tuning using the TQA dataset and a final step of ensemble learning, as seen in the works of ISAAC [7], WSTQ [17]m and MRHF [18]. ISAAC attempts to overcome critical challenges such as the complexity and relatively small size of TQA dataset and the scarcity of large diagram datasets. ISAAC deals with every type of question separately ignoring the correlation between different types of questions and relays on fine-tuning large pre-trained models, ensemble learning, and large datasets. They incorporate a pre-trained transformer (RoBERTa) model to encode text and bottom-up top-down (BUTD) attention with six model ensembles for feature extraction. They use four knowledge retrievers: information retrieval (IR), next sentence prediction (NSP), nearest neighbors (NN), and diagram retrieval. The textual ISAAC is pre-trained on RACE, ARC-Challenge, and OpenBookQA datasets and fine-tuned on CK12-QA. For a visual understanding, they extract the visual features of the diagram constituents and apply BUTD attention to answer diagram questions. They used BUTD highlight the most relevant diagrams or visuals for each question. The multimodal ISAAC is pre-trained on VQA abstract scenes, VQA and AI2D datasets and fine-tuned on CK12-QA.

WSTQ tries to learn effective diagram representations. They applied a text-matching task to comprehend the text and applied a relation-detection task to learn the diagram semantic. They consider the region representations and the relationships between them to learn more effective diagram representation.

MRHF proposed a Multistage Retrieval and Hierarchical Fusion, a novel framework that incorporates dense passage re-ranking and a mixture-of-experts architecture to enhance TQA performance. The framework then integrates multimodal feature fusion to process both textual and heterogeneous diagram features, using a unified question solver for different question types.

The third category uses span-level evidence during the answering process in order to provide an explanation for question answering which — to some extent — achieves a sufficient level of interpretability as in the work of XTQA [17]. XTQA puts explainability first place by providing the students with the explanations accurately which helps them have a deeper understanding of what they have learned. They regard the whole textual context of the lesson as candidate evidence and then applies a fine-grained algorithm to extract span level explanations for answering questions. They apply a self-supervised learning method SimCLR to learn the representation in the TQA dataset.

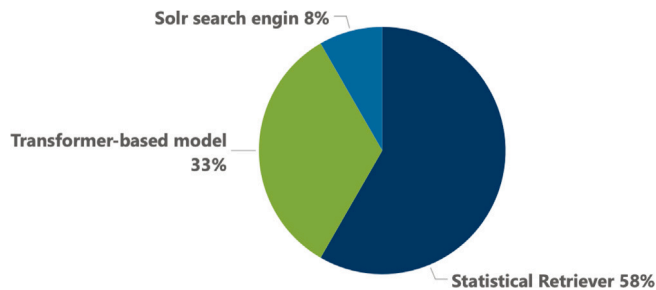

**Fig. 2.** Kind of retrievers used in the current works of TQA task.

While our work also aims to improve TQA task, we focus on refining the textual component of the task as a foundation for future multimodal integration. Unlike previous works, our approach does not yet incorporate visual data. However, by integrating Retrieval-Augmented Generation (RAG) with Llama-2, we address critical challenges such as long contexts, out-of-domain scenarios, and noise reduction in the retrieval process, which are often overlooked in multimodal approaches.

The retrievers in the previous studies, which are an essential component of traditional QA systems, are employed to retrieve relevant passages likely to contain the correct answer. Retrievers can be sparse, relying on classical information retrieval (IR) methods like TF-IDF [19], dense, incorporating DL retrieval methods as in REALM [20] and ORQA [21], or iterative, as in MUPPET [22]. As handling the long context was a main challenge in TQA, efforts to enhance the process of retrieving the relevant text to a question are needed. The retrieved document is then processed through a post-processing or ranking component.

Many current works have used a statistical retriever (TF-IDF) as in the work of MemN, IGMN, EAMB. Some have utilized a search engine as in MHTQA, while others employed transformer-based models (BERT, RoBERTa) as shown in Fig. 2.

Advancements in retrieving information related to questions for comprehension have been made with the use of semantic search [23]. This involves embedding the entire context into a vector space, calculating distances to find measurable relationships, and retrieving the closest (related) ones to an embedded question or query. As TQA retrievers needs to be enhanced with more accurate ways of retrieving relevant context, semantic search serves this need by embedding the entire lessons into a vector space and calculating distances, which improves the comprehension of related context.

Despite significant progress in current TQA systems, a fundamental challenge remains related to reasoning capabilities, text understanding, and handling long contexts within these models, as indicated by the moderately low accuracy achieved. Those models may provide a general answer without going into much detail about the particular responses or results. This suggests that the model's ability to reason in situations requiring a deeper level of comprehension is limited.

## 3. Textbook question answering dataset

The TQA dataset consists of 1076 lessons that cover life science, earth science, and physical science. Each lesson in the TQA dataset is accompanied by multiple-choice questions, with each question offering 2–8 answer options.

The questions are categorized as non-diagram true/false (NDTF) questions, which have a true or false possible answer, non-diagram multiple-choice (NDMC) questions, which have 4–7 answer choices, and diagram multiple-choice (DMC) questions, which have four candidate answers. The dataset focuses exclusively on factoid questions, which are questions that can be answered with simple facts or named entities. The answers will be provided in a multiple-choice form. The answers are provided in a multiple-choice format. The dataset was divided into training, validation, and testing sets, each containing lessons and both non-diagram and diagram questions. Table 1 shows the distribution of all question types and their respective counts, highlighting the total number of ND questions in each split, which are the primary focus of our training and evaluation.

**Table 1**
TQA dataset (CK12-TQA) statistics, including non-diagram (ND) questions.

| Dataset | Train | Dev | Test | Total |
| --- | --- | --- | --- | --- |
| CK12-TQA | 15,154 | 5309 | 5797 | 26,260 |
| – $NDTF$ | 3490 | 998 | 912 | 5400 |
| – $NDMC$ | 5163 | 1530 | 1600 | 8293 |
| – $DMCQuestions$ | 6501 | 2781 | 3285 | 12,567 |
| **ND questions Total** | **8653** | **2528** | **2512** | **13,693** |

## 4. Methodology

### 4.1. Information retrieval

As in the original work of RAG [24], it has been applied to retrieve relevant knowledge to be fed into a generative model like GPT to generate responses with regard to that knowledge.

While RAG was primarily used in open-domain QA, they apply it to retrieve knowledge from external resources such as Wikipedia. In contrast, our work focuses on building a RAG model for the textbook content only to retrieve a domain-specific context where lessons are long and the context is often spread across multiple sections, introducing more complexity. Additionally, we extend RAG by integrating it with a fine-tuned LLM on our specific dataset, Llama-2, to enhance reasoning by generating more contextually relevant answers and benefit from its extensive pre-training.

In the Textbook Question Answering (TQA) task, it is crucial to provide relevant context from the textbook to accurately answer questions. However, the pre-trained large language model (LLM) may not fully capture all the necessary information from the textbook during its pre-training phase, leading to gaps in knowledge. To bridge this gap, our approach involves augmenting the LLM's context window with relevant lesson content retrieved directly from the textbook using Retrieval-Augmented Generation (RAG).

Our experiments demonstrate that relying solely on the LLM's pretrained knowledge is often insufficient, underscoring the importance of RAG in enhancing answer generation. As illustrated in Fig. 3, RAG augments the LLM's knowledge by incorporating the relevant topics from the TQA dataset, effectively combining the pretrained knowledge of the LLM with the domain-specific content from our textbook corpus. This augmentation ensures that the LLM has access to the most relative information when generating answers. For example, when answering a question that requires a context from different lessons, RAG retrieves these relevant parts and feeds them into the LLM's context window, allowing the LLM to produce accurate and contextually relevant answers.

RAG presents a solution to address the issue of scattered information across lessons in TQA, while also reducing the risk of hallucinations in LLM responses by providing contextual grounding for the LLM to infer answers.

The IR pipeline in our architecture, as shown in Fig. 4, begins with vector embedding of the textbook content using the OpenAI text-embedding-ada-002 model. These embeddings are then stored in a vector database, which acts as a searchable repository for retrieving relevant information. When a question is posed, it is first embedded into the same vector space, allowing the query retrieval process to search the database for passages with the highest similarity scores. If necessary, the retrieved information is re-ranked by the reranker [25] to determine its relevance to the query more accurately. Finally, the retrieved passages are added to the LLM's context window, providing

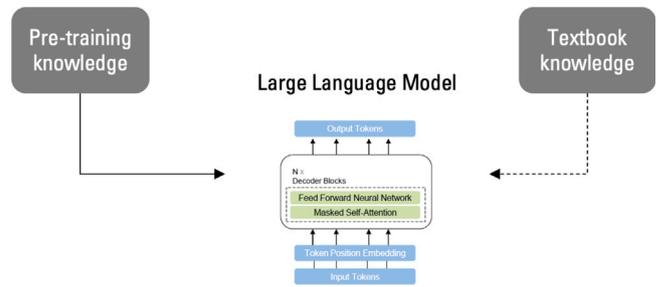

**Fig. 3.** Integrating RAG with LLM. RAG augments the LLM's pretrained knowledge with the textbook knowledge, ensuring a more comprehensive understanding of the textbook content.

the model with relevant information for generating more accurate answers.

The dimensionality of the vector database is determined by the embedding model, and similarity metrics such as cosine similarity or dot product are used to calculate relevance [2], ensuring that the most contextually appropriate passages are retrieved.

The workflow of the RAG in TQA is illustrated in 5. It starts with a TQA question, which may include an image. This question is embedded using an embedding model (text-embedding-ada-002), transforming both the text and image into vector representations. These vectors are stored in a Vector Database (Vector DB) alongside embedded lessons or topics. A retriever then accesses this database to fetch relevant context based on the embedded question. Finally, this retrieved context, along with the original question and answer options, is provided to a large language model (LLM) as a prompt to generate the final answer output. LLMs that incorporate RAG in their architecture are referred to as RAG models [24], and these models have been shown to improve accuracy [26,27].

Our method integrates a RAG framework to accurately and reliably answer scientific questions, particularly addressing the "out-of-domain" problem where some questions require inferring answers from sentences in different lessons. To formulate the problem, we have several questions with Qi representing the vector embedding of question i within the lesson, and several topics from which answers may be derived, with Tj as the vector embedding of topic j. The vector database, created by the search tool, returns relevant chunks through vector search. We use a dot product metric to compute similarity by multiplying the two vectors and measuring the distance between them in terms of their directions. Qi and Tj in Eq. (1) represent the query vector and topic vector, respectively. This similarity score enables the retrieval system to identify the most contextually relevant passages, which are integrated into the LLM's context window to generate accurate answers.

$$\text{Similarity}(Q_i, T_j) = Q_i \cdot T_j \qquad (1)$$

### 4.2. Fine-tuning large language model, Llama-2

The latest language models introduced in the TQA task include pretrained transformers [7,28], aiming to enhance TQA system performance in terms of accuracy. With the growing impact of LLMs on the field of AI, integrating these models into the TQA task becomes crucial for accuracy improvement. Auto-regressive models, or decoder-only models, represent a category of Transformer models trained extensively in a self-supervised manner [29]. While initially designed for predicting the next token in a sequence, they are adapted for QA through techniques like supervised fine-tuning (SFT) and reinforcement learning from human feedback.

Our work utilizes the Llama-2 [9] model as the foundation for fine-tuning due to its significant language understanding and generation capabilities. Llama-2 represents an enhancement over Llama-1 with increased training data and a larger context window (4096 tokens). Trained on 2 trillion tokens of data and scaled to 70 billion parameters, Llama-2 stands out as one of the best-performing open-source LLMs. Its availability for research and prowess in reasoning and reading comprehension make it a valuable asset [9]. Integrating Llama-2's knowledge with the results of RAG adds further depth to its existing training.

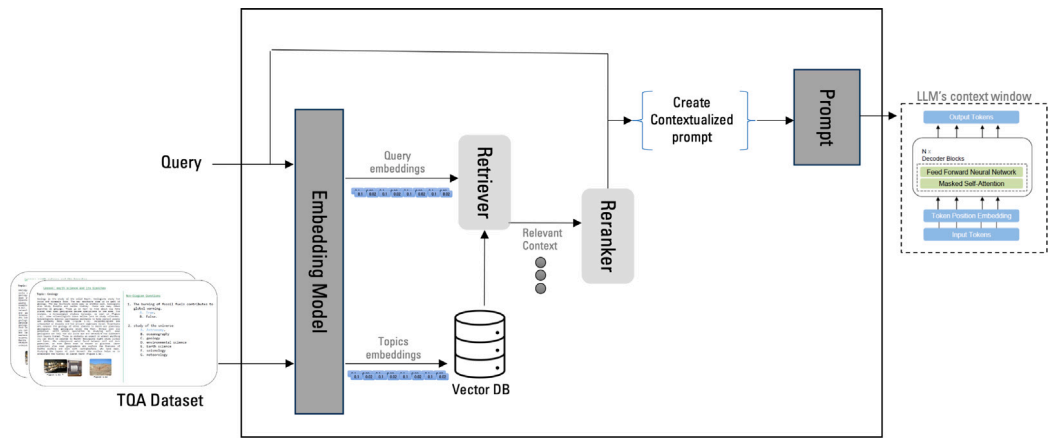

**Fig. 4.** Pipeline of RAG model in TQA task.

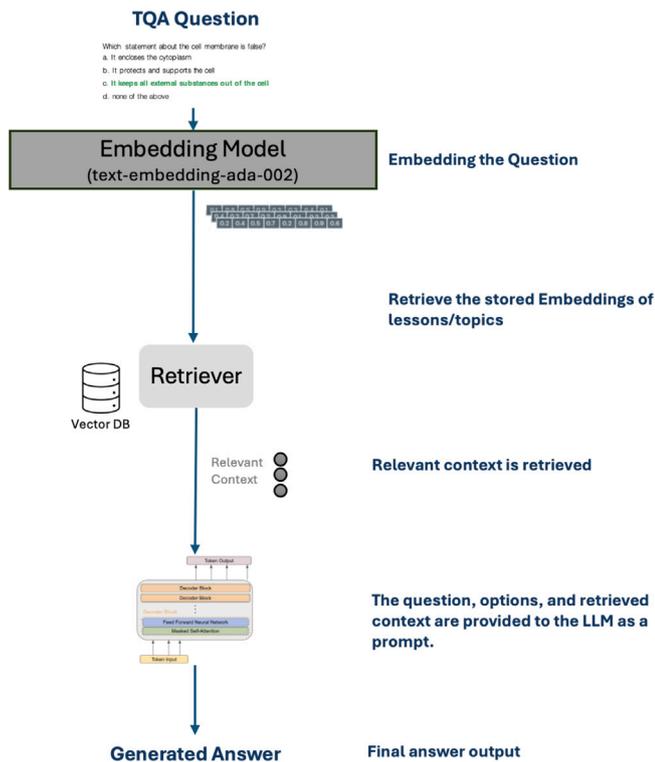

**Fig. 5.** The integration process of Retrieval-Augmented Generation (RAG) with LLaMA-2 for answering textual TQA questions.

Fine-tuning and prompt engineering are employed to enhance LLMs' reasoning abilities [30]. Prompt engineering enables in-context learning via prompts [31,32]. In SFT, models are trained using input examples and corresponding outputs. Prompts guide the model in a way that is optimal for the downstream task.

The process of generating answers for the TQA dataset using RAG approach integrated with the Llama-2 model is outlined in Algorithm 1. The algorithm starts with the preprocessing of the dataset, where text is tokenized and the features are extracted using the embedding model to prepare the data for the model training. Next step focuses on retrieving the top $N$ relevant chunks for every question using the embeddings. The next step is fine-tuning Llama-2 with the retrieved context where we combine the input question, retrieved context, multiple-choice options, and correct answer to fine-tune the LLaMA-2 model. This allows the model to learn how to generate answers using both the retrieved data and labeled examples. The fine-tuned model produces the final answer using the integrated context.

In this algorithm, finetuning refers to adapting the pretrained LLaMA-2 model on the CK12-QA dataset using Supervised Fine-Tuning (SFT). Apply LoRA means adding low-rank adapter layers to the frozen pretrained model weights, which enables efficient training without modifying the base model parameters.

---

**Algorithm 1** Textbook Question Answering Using RAG with LLaMA-2

**Input:** TQA Dataset (CK12-QA) containing questions, diagrams, and multiple-choice options
**Output:** Generated answers for the questions

1 **1. Data Preprocessing**
2   **foreach** *lesson in the dataset* **do**
3     Tokenize all text Extract features using text-embedding-ada-002 model
4   Store preprocessed data in a structured format

5 **2. Retrieving Relevant Context**
6   **foreach** *question in the dataset* **do**
7     Embed the question using the text-embedding-ada-002 model. Retrieve the top N most relevant context chunks (text information) from the vector database based on similarity scores

8 **3. Fine-Tuning LLaMA-2 with Retrieved Context**
9   **Parameter-Efficient Fine-Tuning (PEFT)**:
10   Apply LoRA to add adapter layers to the model
11   Initialize the pre-trained LLaMA-2 model
12   **Supervised Fine-Tuning (SFT)**: **foreach** *(question, retrieved context, options, correct answer) tuple* **do**
13     Combine the question, retrieved context, multiple-choice options, and correct answer
14     Fine-tune only the adapter layers while keeping the main model weights fixed

15 **4. Generating the Final Answer**
16   **foreach** *question* **do**
17     Use the fine-tuned LLaMA-2 model with integrated RAG to generate the final answer.

18 **return** *Generated answers for all questions in the dataset*

```
[INST]<<SYS>>\nYou are a helpful, respectful and honest
assistant. Always answer as helpfully as possible using the
context text provided. Your answers should only be the choice
from the given multiple Options and not have any text after
the answer is done.\n<</SYS>>\n\nContext:
{context}\nQuestion: {question}\nOptions:{options}\nAnswer:
[/INST]{answer}</s>
```

Fig. 6. Prompt format used to fine-tune Llama-2.

Fig. 6 illustrates the general format in which we processed the TQA dataset for the Llama-2 prompt, ensuring compatibility with the CK12-QA dataset and the model architecture.

Llama-2 underwent fine-tuning through transfer learning on the CK12-QA dataset. This phase involved updating the model's weights through additional training epochs specifically designed for the textbook QA domain. The objective of fine-tuning Llama-2 was to improve its performance and safety, thus paving the way for more responsible LLM creation.

The fine-tuning step is crucial for adapting the LLM to a specific task. However, one of the risks associated with fine-tuning large-scale pretrained language models is catastrophic forgetting [33], where the model may forget some of its previous knowledge while updating parameters during fine-tuning. To address this issue, SFT is employed [34].

SFT is tailored to enhance pretrained models for supervised learning tasks using smaller datasets compared to those used for the initial pre-training of the LLM. It offers memory-efficient training using techniques like parameter-efficient fine-tuning (PEFT) [35] to reduce training memory usage and enable the quantization of the model. Quantization involves transforming activations and parameters from floating-point values into lower-precision data types, such as 8-bit integers or even lower [36].

PEFT involves freezing the parameters of the LLM llama2, introducing a trainable layer (LoRA Adapter Layers), and enabling learning only on the newly added layers using examples from our CK12-QA dataset as shown in our finetuning pipeline Fig. 7. Freezing the LLM's parameters and updating only the new parameters on the added layer (adapter layers) brings multiple benefits, including reducing storage requirements, minimizing fine-tuning time, and preventing the loss of LLM knowledge that may occur by updating all model parameters during fine-tuning (see Fig. 7).

## 5. Experiment and results

In this section, We explore the empirical assessment of our suggested methodology, providing a thorough description of the experimental setup, tests that were conducted and results that were attained.

### 5.1. Experimental settings

We conducted experiments that explored the interaction between fine-tuning and RAG to show their effectiveness across the textual questions in the TQA dataset. We first established a baseline performance using zero-shot inference with pretrained LLaMA-2 on CK12-QA. Next, we compared fine-tuning strategies, including supervised fine-tuning (SFT) and parameter-efficient LoRA. For retrieval mechanisms, we integrated RAG with and without re-ranking to assess its impact on accuracy. We also performed subset evaluations on True/False and Multiple-Choice questions, excluding diagram-based content for a fair comparison. Finally, an ablation study analyzed the individual contributions of RAG, finetuning, and re-ranking. This evaluation provides a detailed understanding of our system's performance.

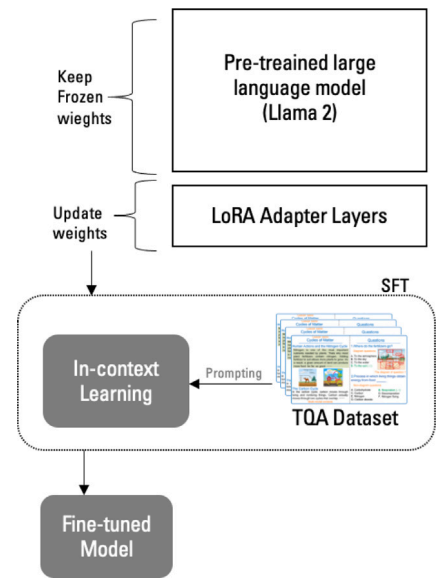

Fig. 7. Our finetuning pipeline.

All training and evaluation were conducted on a single server equipped with an A100 GPU. During the SFT technique, we used SFT-Trainer [34]. We enhanced the parametric-memory generation model Llama-2 with a non-parametric memory through RAG. For the fine-tuning step, we employed a PEFT technique called low-rank adaptation (LORA) [37]. The goal of PEFT, specifically LORA, is to reduce the training cost of Large Language Models (LLMs) with a vast number of parameters, reaching into billions. This is achieved by minimizing the number of updated parameters during fine-tuning. LORA introduces trainable rank decomposition matrices to every layer in the transformer architecture [38]. It adapts only the weight matrices in the self-attention module for downstream tasks and utilizes Adam for model optimization. All weight tensors are quantized to 4 bits using the bitsandbytes library [36] with a method called quantized LORA (QLORA) [39], aiming to decrease the model's size and speed up inference. The primary objective of quantization is to minimize the impact on a model's performance while reducing its memory footprint and processing requirements.

In the SFT stage, we used processed CK12-QA data with a cosine learning rate schedule (learning rate: $2\times10^{-4}$), a weight decay of 0.001, a batch size of 4, and a sequence length of 512 tokens. For the fine-tuning process, each sample consisted of a prompt and an answer. To ensure the model sequence length is properly filled, we concatenated all the prompts and answers from the training set, using a special token to separate prompt and answer segments. We employed an auto-regressive objective, zeroing out the loss on tokens from the user prompt, and backpropagating only on answer tokens. Finally, we fine-tuned the model for 2 epochs. Regarding LORA configurations, we set the Alpha parameter to 16, used a dropout parameter of 0.1, and the rank of the update matrices used for LORA is 64.

**Table 2**

Zero-shot inference accuracy (%) of LLaMA-2 on the CK12-QA dataset's validation and test sets for textual questions (True/False and Multiple Choice).

| Work | Text T/F | Text MC | Text all |
|---|---|---|---|
| LLaMA-2 (Zero-Shot Validation) | 46.09 | 27.39 | 34.77 |
| LLaMA-2 (Zero-Shot Test) | 48.14 | 27.39 | 35.87 |

**Table 3**

Experimental results (accuracy %) of TQA approaches on textual questions (true/false and multiple choice) on validation and test sets.

| Work/Split | Year | Validation (%) | Test (%) |
|---|---|---|---|
| MemN [4] | 2017 | 38.83 | – |
| IGMN [6] | 2018 | 46.88 | – |
| EAMB [41] | 2018 | 41.97 | – |
| F-GCN [3] | 2018 | 54.75 | – |
| ISAAQ [7] | 2020 | 71.76 | 72.13 |
| XTQA [17] | 2020 | 41.32 | 41.67 |
| MHTQA [42] | 2021 | 74.61 | – |
| WSTQ [28] | 2021 | 64.87 | 65.15 |
| MoCA [14] | 2021 | **78.28** | – |
| RAFR [15] | 2021 | 43.35 | 41.03 |
| SSCGN [13] | 2022 | 76.10 | **74.40** |
| DDGNet [43] | 2023 | 41.62 | 41.96 |
| MRHF [18] | 2024 | – | 82.48 |
| PLRTQA$_{Proposed}$ | 2024 | **82.40** | **84.24** |

**Table 4**

Experimental results (accuracy %) of TQA approaches on textual questions (true/false and multiple choice) on validation set.

| Work | Text T/F | Text MC | Text All |
|---|---|---|---|
| MemN | 50.4 | 22.7 | 36.55 |
| IGMN | 57.41 | 40.0 | 46.88 |
| EAMB | 55.31 | 33.27 | 41.97 |
| F-GCN | 62.73 | 49.54 | 54.75 |
| ISAAC | 81.36 | 71.11 | 75.16 |
| XTQA | 58.24 | 30.33 | 41.32 |
| MHTQA | 82.87 | 69.22 | 74.61 |
| WSTQ | 76.65 | 56.30 | 64.33 |
| MoCA | 81.56 | 76.14 | 78.28 |
| RAFR | 53.63 | 36.67 | 43.35 |
| SSCGN | 81.50 | 72.58 | 76.10 |
| DDGNet | 57.74 | 31.15 | 41.62 |
| DIMP | 74.07 | 54.55 | 61.24 |
| MRHF | **87.88** | 78.95 | **82.48** |
| PLRTQA$_{Proposed}$ | 80.72 | **84.97** | 82.40 |

**Table 5**

Experimental results (accuracy %) of TQA approaches on textual questions (true/false and multiple choice) on test set.

| Work | Text T/F | Text MC | Text All |
|---|---|---|---|
| MemN | – | – | – |
| IGMN | – | – | – |
| EAMB | – | – | – |
| F-GCN | – | – | – |
| ISAAC | 77.74 | 68.94 | 72.13 |
| XTQA | 56.22 | 33.40 | 41.67 |
| MHTQA | – | – | – |
| WSTQ | 77.28 | 58.27 | 65.15 |
| MoCA | 81.36 | 76.31 | 78.14 |
| RAFR | 52.75 | 34.38 | 41.03 |
| SSCGN | 79.16 | 71.68 | 74.40 |
| DDGNet | 57.28 | 33.25 | 41.96 |
| DIMP | 75.58 | 55.22 | 62.61 |
| MRHF | **86.62** | 80.00 | 82.40 |
| PLRTQA$_{Proposed}$ | 83.50 | **85.53** | **84.24** |

The RAG was implemented at the paragraph level by treating every topic within the CK12-QA lessons as an independent chunk. Embeddings for these topics were generated using OpenAI's text-embedding-ada-002 model and stored in a scalable Pinecone vector database. The same embedding model was used during inference to retrieve top-k relevant paragraphs via dot-product similarity. We have used Cohere's re-ranking model rerank-english-v2.0 to refine the retrieved results further.

*5.2. Main result*

We conducted a zero-shot inference experiment on the CK12-QA dataset using the pre-trained model llama2 without finetuning on all non-diagram diagram questions including both multiple choice and true/false format. As depicted in Table 2 The overall accuracy on the validation set is 34.77% with 46.09 for T/F questions and 27.39% for MC questions. On the test set, the overall accuracy achieved is 35.87% with 48.14 for T/F questions and 27.39% for MC questions. The low zero-shot performance results and the struggle with MC questions compared to the simpler format of T/F questions show that the model cannot handle complex reasoning and long lesson comprehension in TQA tasks without further training. This performance gap compared to our fine-tuned, RAG-augmented approach highlights the importance of TQA task adaptation and retrieval-based enhancements.

The evaluation of the fine-tuned language model and the RAG technique was conducted on the CK12-QA dataset. The fine-tuning of Llama-2 was carried out on the training set following the SFT methodology.

Inspired by [40], we fine-tuned Llama-2 on two categories of non-diagram questions: multiple choice and true/false questions simultaneously. The model was trained to complete the given prompt, and the loss was optimized based on the provided answer by the model. Instead of penalizing the model for predicting the next token in the given prompt, we passed a response template to the collator. This adjustment allowed the model's weights to be adjusted based on the answers it provided for the questions.

In this approach, the context in the model's prompt was the complete lesson of the question to which it belongs. This context improvement in Llama-2 resulted in an enhancement in the test set performance. Table 3 illustrates a 1.41% decrease in accuracy scores on the validation set, achieving an accuracy of 82.40%. However, the accuracy increased by 0.525% on the test set for all text questions, reaching an accuracy of 84.24%. This demonstrates the effectiveness of providing the entire lesson as context to the LLM.

The performance of the fine-tuned model was assessed using accuracy and compared with related works for every category of questions in TQA, as shown in Tables 4 and 5. Utilizing comprehensive context significantly improved accuracy, and the use of RAG in the entire lesson resulted in different trade-offs between validation and test set accuracies. Our best model achieved an improvement of 5.02% on the test set overall accuracy and delivered the best performance on MC questions across both validation and test sets, outperforming previous state-of-the-art methods on the TQA task.

Tables 4 and 5 provide a detailed breakdown of our model's performance on textual questions, focusing on True/False (T/F) and Multiple-Choice (MC) questions, with diagram-based content excluded. To ensure fair comparisons, we used the same dataset splits as previous works, focusing on non-diagram questions only. This allows for consistent and reliable evaluation of performance.

Table 4 provides a breakdown of results on the validation set. The proposed model achieves an accuracy of 80.72% on T/F questions and 84.97% on MC questions, resulting in an overall accuracy of 82.40%. However, MRHF performs best in T/F questions with 87.88% and achieves a slightly higher overall accuracy of 82.48%, while our method delivers the best performance on MC questions. These results highlight the effectiveness of our method, particularly for complex MC tasks present in TQA task.

On the test set, presented in Table 5, our model achieves the best performance in MC questions with an accuracy of 85.53%, outperforming MRHF's 80.00%. For T/F questions, MRHF performs better with 86.62%, compared to our model's 83.50%. Our work achieves the highest overall accuracy on the test set with 84.24%, outperforming MRHF's 82.40%. These results prove the strength of our approach in handling challenging MC questions while maintaining competitive performance across all categories.

Table 6
Validation accuracy scores for different model configurations.

| Configuration | Validation accuracy (%) |
| --- | --- |
| LLaMA-2 without fine-tuning | 34.77 |
| LLaMA-2 with full lesson context (No RAG) | 82.40 |
| LLaMA-2 with RAG (No Re-ranker) | 83.58 |
| LLaMA-2 with RAG and full lesson context | 83.78 |
| LLaMA-2 with RAG + Re-ranker | 80.74 |

Table 7
Test accuracy scores for different model configurations.

| Configuration | Test accuracy (%) |
| --- | --- |
| LLaMA-2 without fine-tuning | 35.87 |
| LLaMA-2 with full lesson context (No RAG) | 84.24 |
| LLaMA-2 with RAG (No Re-ranker) | 83.80 |
| LLaMA-2 with RAG and full lesson context | 81.73 |
| LLaMA-2 with RAG + Re-ranker | 79.22 |

*5.3. Ablation study*

By this section, we aim to further understand the impact of RAG component on our finetuned model. Before incorporating RAG into our model, we began by analyzing the number of questions in the training data that require context from lessons other than the ones they belong to. We measured the similarity between the questions and the topics inside lessons by calculating the dot product of the vector-based topics stored in a vector database and the question's vector. Out of the 8653 samples in the training dataset of non-diagram questions, we found that only 44% of the questions obtained a retrieved topic from the lesson they belong to, leaving the rest with topics from other lessons.

When incorporating a re-ranking mechanism, the percentage of questions that obtain answers from the lesson they are in rises to 52.43%. This analysis shows us the importance of using techniques that solve the "out-of-domain" problem by inferring the answers from context outside the lesson they belong to.

The evaluation results of the fine-tuned language model on the TQA dataset are presented in Tables 6 and 7. The performance evaluation of the Llama-2 model without fine-tuning on CK12-QA examples revealed accuracy scores of 34.77% and 35.87% on the validation and test sets for all text questions, respectively. All text questions includes the non-diagram T/F questions and non-diagram MC questions. This low zero-shot accuracy emphasizes the need for fine-tuning LLMs on domain-specific datasets and suggests the model's limited capability in accurately answering CK12-QA questions.

While our experiments demonstrate the effectiveness of fine-tuning LLaMA-2 using LoRA, we recognize the importance of evaluating LoRA hyperparameters such as rank and dropout to further optimize performance, we leave this exploration for future work.

When RAG is integrated without the re-ranker module, the accuracy notably increased to 83.58% on the validation set and 83.80% on the test set. This highlights the significant impact of including RAG on the model's generative capability without using the re-ranker module.

By omitting RAG, the context in the model's prompt will be the complete lesson of the question that it belongs to. This improvement is demonstrated in the test set results. When both the entire lesson and the retrieved topics are incorporated as context to our model using RAG, an accuracy of 83.78% was achieved on the validation set. However, a slightly lower score of 81.73% was observed on the test set. This suggests a minor trade-off in generative performance when increasing the context window size to a limit that may reach or exceed 4096, which is the maximum window size of Llama-2.

Incorporating the re-ranked relative topics using RAG within the entire lesson as the context for the model decreased accuracy in both the validation and test sets. The validation set achieved an accuracy of 80.74%, while the test set obtained an accuracy of 79.22% for all text questions. This could be a result of feeding the model with unrelated topics that distract it and worsen the inference process of the model.

We used in this work text-embedding-ada-002 model for retrieval as it provides balance between accuracy and computational cost. Future work could explore alternative embedding models and advanced re-ranking strategies to further enhance retrieval quality and its interaction with finetuning.

*5.4. Case study*

We conducted a qualitative analysis to explore how the addition of RAG with the fine-tuned LLama-2 improved the model's performance. This analysis highlights both the successes and limitations of the model when it retrieves information from lessons other than the one the question belongs to. The integration of RAG allowed the model to retrieve relevant context for many questions. We selected two examples from the CK12-QA dataset to illustrate both the strengths and limitations of RAG in retrieving relevant information. These examples show how RAG sometimes succeeded in retrieving the correct context, and other times failed to provide adequate information for the model to make the correct inference.

One successful example can be seen in Fig. 8A. In this example, the model was asked to identify which of the listed options is not caused by gravity in relation to erosion. The retrieved context discusses how "Weathering wears rocks at the Earth's surface due to different agents". This was useful because it reinforced the understanding that while glaciers, flowing water, and landslides are all influenced by gravity, moving air (wind) is not directly driven by gravity in the context of erosion. Hence, the model correctly selected moving air as the exception, with the context helping to clarify the relationships between the agents of erosion and gravity.

However, RAG also revealed limitations when the retrieved context did not align with the question's requirements, resulting in incorrect answer. An example of this can be seen in Fig. 8B, the model with RAG was asked to identify the factors that increase the risk of landslides, the model incorrectly predicted (D) human activities. The context mentioned "Layers of weak rock, such as clay, also allow slopes to slide", which may have led the model to focus more on human-induced alterations of weak rock layers, rather than recognizing the role of earthquakes. In this case, the context distracted the model from identifying the correct answer, resulting in a misprediction. To gain a deeper understanding of how RAG handled retrieval across different lessons, we analyzed cases where the model retrieved context from a lesson other than the one the question originated from.

As shown in Fig. 8C, For the question (NDQ_018937), the model successfully answered the question by retrieving relevant information from a different lesson than the one it is in, which ultimately helped it select the correct answer. This highlights RAG's ability to synthesize information across lessons and apply it effectively, which is very important in educational contexts where overlapping concepts between lessons are common. This capability is especially useful in educational environments where overlapping concepts are spread across different lessons.

**6. Conclusion**

Textbook question answering (TQA) poses a significant challenge in the field of artificial intelligence (AI) due to the complexity it brought. The field is evolving rapidly with the introduction of new large language models (LLMs), fundamentally changing the dynamics of the

(A) Success case: The context was helpful, as it clarified the role of gravity in erosion and guided the model with RAG to the correct answer.

(B) Fail case: The context distracted the model with RAG from identifying earthquakes as the correct answer, leading to confusion and an incorrect prediction.

(C) Success case: The retrieved context from different lessons clarified the relationship between force and work, helping the model answer the question correctly.

**Fig. 8.** Case studies illustrating the effectiveness of context retrieval with RAG.

game. This paper contributes to the advancement of TQA by focusing on the textual challenges and leaving the integration of the visual part for future work. We rely on recent trends in transfer learning with the goal of enhancing the reasoning capabilities of TQA systems and, consequently, improving overall accuracy. Leveraging the immense potential of LLMs, we specifically adapt the LLM model Llama-2 to the TQA task through SFT. Additionally, we incorporate the RAG technique to enhance the quality of text generated by the LLM and tackle the "out-of-domain" problem. One of the primary strengths of our work is the successful adaptation of LLaMA-2 to the TQA task, which led to a significant improvement in answering text-based questions. Specifically, our architecture achieved an accuracy improvement of 4.12% on the validation set and 9.84% on the test set for textual multiple-choice questions, including true/false and non-diagram questions. These results demonstrate the effectiveness of combining RAG with LLMs in tackling the "out-of-domain" problem and ensuring that the model can retrieve relevant information from textbook content to generate accurate answers. Our study not only demonstrates the potential of LLMs, such as LLaMA-2, when paired with retrieval mechanisms like RAG, but also establishes a solid basis for further research in multimodal TQA. While this work contribute to the field of educational AI systems by improving the textual question-answering process on textual TQA as a foundational base, we recognize the need to address more complex multimodal tasks that incorporate visual parts such as diagrams and figures. Handling diagram-based tasks requires additional modifications; therefore, future research will extend our framework to integrate advanced visual understanding techniques that tackle the TQA multimodal challenges.

**CRediT authorship contribution statement**

**Hessa A. Alawwad:** Writing – original draft. **Areej Alhothali:** Supervision. **Usman Naseem:** Supervision. **Ali Alkhathlan:** Supervision. **Amani Jamal:** Supervision.

**Research funding**

Author Hessa Alawwad: This project was funded by the Deanship of Scientific Research (DSR) at King Abdulaziz University, Jeddah, Saudi Arabia, under grant no. (GPIP: 1131-612-2024). The authors, therefore, acknowledge with thanks DSR for technical and financial support.

**Declaration of competing interest**

The first author is employed by Imam Mohammad Ibn Saud Islamic University (IMSIU) and study at King Abdulaziz University, which have no commercial interest in the research outcomes. Remaining authors have no conflict of interest to declare.

The authors affirm that these declared interests have not influenced the design, conduct, analysis, or interpretation of the research presented in this paper. The research was conducted impartially and with the utmost integrity.

**Acknowledgments**

This project was funded by the Deanship of Scientific Research (DSR) at King Abdulaziz University, Jeddah, under grant no. (GPIP: 1131-612-2024). The authors, therefore, acknowledge with thanks DSR for technical and financial support.

**Data availability**

The data used was described in detail in the paper.

**Hessa Abdulrahman AlAwwad** specialize in the field of Computer Science, with a keen interest in cutting-edge technologies such as Artificial Intelligence (AI), Natural Language Processing (NLP), and Computer Vision (CV). Her current passion lies in the exciting realm of multimodal question answering, where he explore the intersection of AI, NLP, and CV to tackle complex challenges in understanding and responding to questions involving multiple types of data. Her work is driven by a relentless curiosity and a commitment to advancing the frontiers of knowledge in these domains.